\documentclass[10pt,journal,compsoc]{IEEEtran}
\usepackage{pifont}
\usepackage{balance}
\usepackage{enumitem}
\usepackage{graphicx}
\usepackage{subfigure}
\usepackage{multirow}
\usepackage{tablefootnote}
\usepackage{threeparttable}
\usepackage[ruled,vlined,linesnumbered,ruled]{algorithm2e}
\usepackage{color}
\usepackage{algorithm2e}
\usepackage[noend]{algpseudocode}
\usepackage{xspace}
\usepackage{amsmath}

\usepackage{amsfonts}
\usepackage{url}
\usepackage{makecell}
\usepackage{diagbox}
\usepackage{booktabs}
\usepackage{cite}

\usepackage{caption}

\usepackage{fancyhdr}

\fancypagestyle{normalstyle}{%
\fancyfoot[C]{\fontsize{10}{30}\selectfont \thepage} 
}
\usepackage{caption}
\usepackage{wasysym}
\usepackage{tikz}
\newcommand*{\circled}[1]{\lower.8ex\hbox{\tikz\draw (0pt, 0pt)%
    circle (.47em) node {\makebox[0.4em][c]{\small #1}};}}

\def\ie{\textit{i.e.}\xspace}
\def\vs{\textit{vs.}\xspace}
\def\etal{\textit{et al.}\xspace}
\def\etc{\textit{etc.}\xspace}

\def\eg{\textit{e.g.}\xspace}

\def\method{HAT\xspace}
\def\prefill{\textit{prefill}\xspace}
\def\decode{\textit{decode}\xspace}

\begin{document}



\title{A Novel Hat-Shaped Device-Cloud Collaborative Inference Framework for Large Language Models}

\author{
        Zuan Xie,~
 	Yang Xu,~\IEEEmembership{Member,~IEEE,}~
 	Hongli Xu,~\IEEEmembership{Member,~IEEE,}~
        Yunming Liao,~ 
        Zhiwei Yao~
 	\IEEEcompsocitemizethanks{
 		\IEEEcompsocthanksitem Z. Xie, Y. Xu, H. Xu, Y. Liao and Z. Yao are with the School of Computer Science and Technology, University of Science and Technology of China, Hefei, Anhui, China, 230027, and also with Suzhou Institute for Advanced Research, University of Science and Technology of China, Suzhou, Jiangsu, China, 215123. E-mails: xz1314@mail.ustc.edu.cn; xuyangcs@ustc.edu.cn; xuhongli@ustc.edu.cn; ymliao98@mail.ustc.edu.cn; zhiweiyao@mail.ustc.edu.cn.
   }
}


\markboth{}{Z. Xie \MakeLowercase{\textit{et al.}}: HAT: A Novel Device-Cloud Collaborative Inference Framework for Large Language Models}

\IEEEtitleabstractindextext{%
\begin{abstract}
Recent advancements in large language models (LLMs) have catalyzed a substantial surge in demand for LLM services. 
While traditional cloud-based LLM services satisfy high-accuracy requirements, they fall short in meeting critical demands for low delay and enhanced privacy. 
To address these limitations, we propose \method, a novel device-cloud collaborative inference framework that leverages the complementary strengths of U-shaped inference and speculative decoding.
\method partitions the LLM into three submodels, and the input and output submodels, stacked with a lightweight adapter network, are deployed as a small language model (SLM) on each end device. 
Meanwhile, the middle submodel, encompassing the majority of the LLM's decoder layers, is hosted in the cloud to perform speculative decoding with on-device SLMs.
During inference, \method exchanges hidden states (rather than raw tokens) of input or draft tokens between devices and the cloud, thereby incurring substantial communication delays.
Besides, processing hidden states of long prompts will exacerbate computation delays in the cloud, further compromising inference efficiency.
To improve efficiency, we introduce a prompt chunking mechanism that segments long prompts into shorter chunks, enabling parallel transmission and processing. 
Furthermore, \method is implemented to dynamically determine optimal chunk sizes for devices handling long prompts, thereby improving overall inference speed.
Extensive experiments are conducted on a physical testbed comprising 30 NVIDIA Jetson devices and a server with 8 NVIDIA A6000 GPUs. 
Experimental results demonstrate that \method achieves promising performance improvements, reducing TTFT by 41\% to 54\% and TBT by 41\% to 77\% compared to the baselines. 

\end{abstract}
}

\maketitle
\IEEEdisplaynontitleabstractindextext
\IEEEpeerreviewmaketitle

\section{Introduction}\label{sec:intro}

Recent advancements in large language models (LLMs) have revolutionized the field of natural language processing, demonstrating unprecedented capabilities across various tasks and triggering exponential growth of LLM services \cite{wang2024survey,li2024pre}. 
For instance, OpenAI's ChatGPT provides various services, \eg, chat-based interaction, and automated writing, to approximately 180 million users, and processes over 1.6 billion requests monthly \cite{duarte2023chatgpt}.
The underlying architecture of LLM services mainly operates through an autoregressive process, which involves a \prefill phase followed by a \decode phase. 
In \prefill phase, the LLM processes all input prompt tokens simultaneously, leveraging parallel computation to generate the initial output token. 
During \decode phase, the LLM generates the rest of the content token by token, where the latest token is processed at each inference step.

Generally, the LLM services are powered by cloud servers with sufficient computing power, and can provide highly accurate solutions to complex requests \cite{zhang2019mark}.
Although cloud-based services satisfy the high accuracy demands of many applications, they fall short in meeting critical requirements like low delay and enhanced privacy. 
Specifically, the potential nature of the autoregressive process significantly reduces the inference speed, as only one token will be generated per inference step.
Furthermore, directly uploading raw data to the cloud for inference exposes sensitive information during transmission or storage, and the inference output may also contain privacy-sensitive information.
It is reported that over 81\% of interviewed users are not satisfied with the cloud-based LLM services regarding the delay and privacy concerns, which are essential for some popular applications, such as autonomous driving and health monitoring \cite{li2024personal,cui2024survey, de2024assessing}.

As the computing capabilities of end devices increase, many studies propose device-cloud collaborative inference frameworks to address the two concerns in cloud-based LLM services \cite{he2024large}.
We summarize the advantages and disadvantages of different inference frameworks in Table \ref{table:method}.
Concretely, a basic collaborative inference framework (as adopted in Apple Intelligence \cite{gunter2024apple}) is proposed to deploy small language models (SLMs) on end devices to handle some simple delay-sensitive or privacy-sensitive requests locally, while other complex requests are offloaded to the in-cloud LLMs for processing. 
With local SLMs, this framework can improve the inference speed and alleviate the privacy issue to a certain extent.
However, the limited performance of SLMs may result in reduced inference accuracy and user experience.


In order to maintain satisfactory inference accuracy while improving overall inference speed, speculative decoding (SD) has been integrated into the device-cloud inference frameworks \cite{hao2024hybrid, oh2024uncertainty, zheng2025citer}.
Following the idea of SD, the SLM on an end device generates a multi-token draft sequence, which undergoes parallel verification by the in-cloud LLM with a single inference step (also called a verification step).
The parallel verification significantly accelerates inference compared to traditional sequential generation while preserving high accuracy through the rejection of divergent draft tokens and subsequent content.
However, as the input and output tokens are shared between devices and the cloud, this framework still inevitably raises significant privacy concerns. 

To enhance privacy, split inference has been proposed as a privacy-preserving device-cloud inference framework \cite{yan2024protecting, mai2023split, ohta2023lambda}.
This framework splits an LLM into two submodels and deploys them separately on end devices and the cloud.
Then, LLM inference is performed in a pipeline fashion, where the intermediate data (\ie, hidden states) of each on-device submodel (rather than raw tokens) are uploaded to the cloud to derive output tokens.
To further protect the privacy of sensitive outputs, a variant of split inference, called U-shaped inference, partitions an LLM into three submodels \cite{sa2024ensuring}. 
The input and output submodels are deployed on devices, while the computationally intensive middle submodel, encompassing the majority of transformer layers, resides in the cloud. 
Although this framework effectively safeguards both input and output tokens from exposure beyond devices, it incurs considerable communication overhead for the exchange of hidden states, which are much larger than tokens \cite{vaswani2017attention}. 
Moreover, during \decode phase, the iterative generation process necessitates frequent transmission of hidden states between devices and the cloud, reducing decoding efficiency.

As far as we know, existing device-cloud inference frameworks have struggled to concurrently optimize delay, privacy, and accuracy.
By leveraging the complementary strengths of U-shaped inference (for privacy) and speculative decoding (for delay), we propose a novel device-cloud collaborative inference framework termed \textit{\method}. 
Similar to the U-shaped framework, \method deploys the LLM's input and output submodels on devices to protect privacy. 
Besides, a lightweight adapter network is constructed via knowledge distillation and combined with the LLM's input and output submodels to form an SLM, called draft model. 
The on-device draft model performs speculative decoding with the LLM, effectively reducing the frequency of device-cloud communication during the inference procedure and maintaining promising accuracy.
However, in \method, devices with long prompts will suffer from significant communication delays when transmitting all hidden states at once during \prefill phase. 
Moreover, since the hidden states of devices operating in different phases (\prefill and \decode) are processed in batch by the LLM simultaneously at each inference step, the devices in \decode phase will experience prolonged in-cloud computation delays when their hidden states are batched with those from the devices with long prompts in \prefill phase.

\begin{table}[t]
\caption{Advantages and Disadvantages of Different Device-Cloud Collaborative Inference Frameworks.}
\label{table:method}
\centering
\begin{tabular}{c|cccc}
\hline
\textbf{Metric}  & \textbf{Basic} & \textbf{SD} & \textbf{U-shape} & \textbf{Ours} \\ 
\hline
Delay     & \LEFTcircle & \CIRCLE & \Circle & \CIRCLE \\
Privacy   & \LEFTcircle & \Circle & \CIRCLE & \CIRCLE \\ 
Accuracy  & \LEFTcircle & \CIRCLE & \CIRCLE & \CIRCLE \\ 
\hline 
\end{tabular}
\end{table}

To address the degradation of communication and computational efficiencies caused by long prompts, we propose parallelizing the transmission and processing of hidden states through prompt chunking. 
During \prefill phase, \method divides long prompts into multiple chunked prompts, enabling the overlapping of device-side transmission of chunked hidden states with in-cloud computation of previously adjacent chunked hidden states, thereby effectively reducing bulk communication delays. 
Moreover, prompt chunking also alleviates in-cloud computation delays for devices in \decode phase, as their hidden states are batched with much shorter chunked hidden states from devices in \prefill phase. 
However, excessively large chunks diminish the benefits of parallelization, while overly small chunks inversely lead to increased computation delays for devices in \prefill phase due to fragmentation, which necessitates optimizing the size of chunks in \method.
Our main contributions are as follows:
\begin{itemize}[topsep=0pt]
\item 
We introduce a novel device-cloud collaborative inference framework, termed \method, that integrates U-shaped inference with speculative decoding to deliver LLM services with promising accuracy, low latency, and enhanced privacy.

\item 
We propose a prompt chunking mechanism to enhance the communication and computational efficiencies of \method, especially in scenarios involving long prompts. 
As the size of prompt chunks has a comprehensive influence on the processing of hidden states for devices in different phases during inference. 
Thus, \method is designed to dynamically determine optimal chunk sizes for devices handling long prompts, thereby accelerating LLM inference.

\item 
The performance of \method is evaluated through a physical platform with a total of 30 NVIDIA Jetson devices. 
The experimental results show that \method can reduce the time-to-first-token (TTFT) by 41\% to 54\%, while reducing the time-between-tokens (TBT) by 41\% to 77\%, compared to the baselines.
\end{itemize}

\section{Background and Motivation}\label{sec:prelim}

\subsection{General Inference Process of LLMs}\label{General Inference Process of LLMs}

\textbf{Autoregressive Inference for A Single Request.}
Popular LLMs, such as GPT-3 \cite{gpt3} and LLaMa \cite{touvron2023llama}, are decoder-only models,
which are built from a stack of decoder layers with identical structures, each consisting of two core components: a self-attention module and a feed-forward neural network \cite{vaswani2017attention}. 
The self-attention module computes context-aware weights for token interactions, allowing the model to capture and aggregate semantic relationships across the entire sequence effectively. 
Subsequently, the feed-forward network processes these relationships to generate updated hidden states, which are then propagated to the next layer for further refinement.

The inference process of LLMs mainly operates in an autoregressive manner, and consists of two distinct phases: \prefill phase and \decode phase. 
In \prefill phase, the LLM processes all tokens of a request's input prompt in parallel to produce the first output token. 
Following this, \decode phase begins to generate the remaining tokens sequentially with one token at a time. 
Concretely, at each inference step, the LLM predicts the next token for the request based on its input prompt and all previously generated tokens. 
This process continues iteratively until an end-of-sequence (EOS) token is produced or the maximum token limit is reached. 
A key requirement of \decode phase is access to the keys and values of all previously processed tokens for attention computations. 
To avoid redundant calculations, modern LLM systems utilize a key-value (KV) cache to store precomputed keys and values, allowing for efficient attention computations during the generation of subsequent tokens.

\textbf{Batched Inference for Multiple Requests.}
Usually, commercial LLM systems are designed to handle multiple concurrent requests from various devices. 
Processing these requests sequentially will result in low throughput, as GPU resources often remain underutilized while waiting for individual requests to complete. 
To improve the throughput, many LLM systems adopt continuous batching \cite{yu2022orca, agrawal2024taming} to process multiple requests within a single batch, which is called \textit{Batched Inference}.
It allows requests to dynamically join or leave the batch after completing each inference step, eliminating the need to wait for all requests in the batch to finish their final generation.

The computational resources for batched inference primarily depend on the total number of tokens in the batch\cite{agrawal2024taming}
Since the KV-cache stores the results of all previously processed tokens, the token size for \decode phase is effectively 1, whereas the token size for \prefill phase corresponds to the length of the input prompt. 
As a result, batching is highly efficient for \decode phase, as it significantly improves GPU utilization with minimal delay. 
However, when processing requests from different phases (\ie, \prefill and \decode) within the same batch, requests with long prompts in \prefill phase always consume a disproportionate amount of computational resources. 
This imbalance leads to increased computation delays for other requests in \decode phase.
To mitigate the issue, recent studies \cite{agrawal2024taming, holmes2024deepspeed} propose to divide the prompts of devices in \prefill phase into many shorter chunks. 
These chunks are processed across multiple inference steps, where the first output token of each device is generated until all its prompt chunks are fed to the LLM.
The detailed evaluation is presented in Sec. \ref{subsec_motivation}.

\begin{figure*}[t]
    \centering
    \subfigure[Delay of different inference frameworks]{
        \includegraphics[width=0.22\textwidth, height=3.6cm]{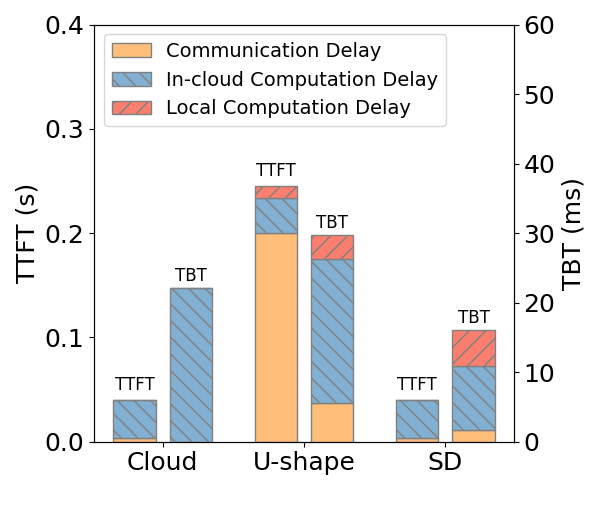}
        \label{fig: decode_time}
    }
    \subfigure[TTFT of U-shaped inference with different prompt lengths]{
        \includegraphics[width=0.22\textwidth, height=3.6cm]{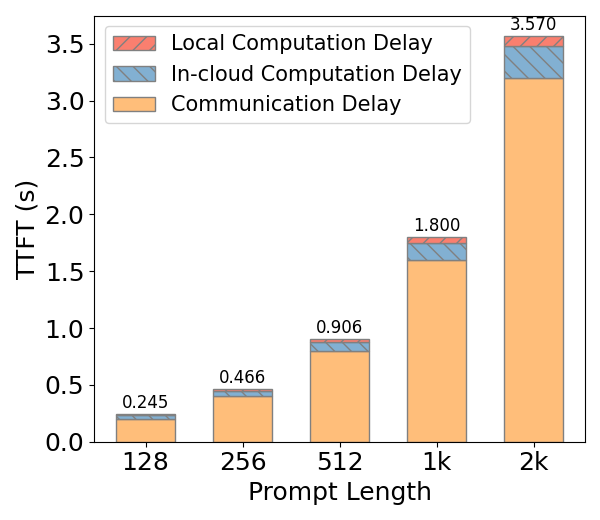}
        \label{fig: Ushaped}
    }
    \subfigure[In-cloud computation delay with different prompt lengths]{
        \includegraphics[width=0.22\textwidth, height=3.6cm]{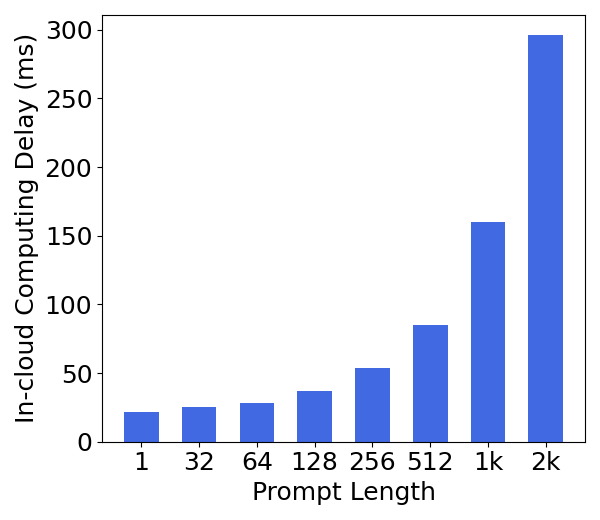}
        \label{fig: batched_token_size}
    }
    \subfigure[Effect of prompt chunking with different chunk sizes]{
        \includegraphics[width=0.24\textwidth, height=3.6cm]{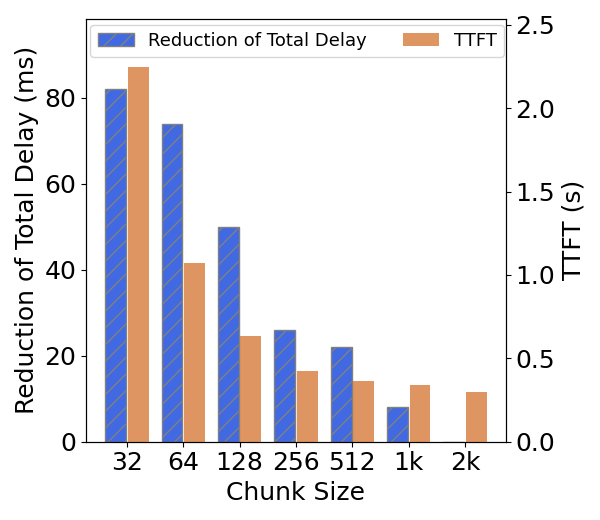}
        \label{fig: chunk_size}
    }
    \vspace{-0.3cm}
    \caption{Results of Preliminary Experiments.}
    \vspace{-0.2cm}
    \label{fig:combined}
\end{figure*}

\subsection{Collaborative Inference}
Our framework integrates the distinct strengths of U-shaped inference and speculative decoding to simultaneously enhance inference latency, data privacy, and accuracy for LLMs.
Below, we provide a detailed explanation of the two collaborative inference frameworks.

\textbf{Speculative Decoding.}
The autoregressive nature of LLMs inherently constrains their inference speed, particularly in cloud-centric LLM systems.
To mitigate this limitation, speculative decoding has been incorporated in the device-cloud collaborative inference framework that employs both an SLM and an LLM for token drafting and verification, respectively \cite{hao2024hybrid, oh2024uncertainty}.
Each round of speculative decoding consists of two stages: drafting and verification.
In the drafting stage, the SLM, typically deployed on end devices, generates a sequence of multiple tokens through autoregressive processing. 
Subsequently, during the verification stage, this draft sequence is fed to the in-cloud LLM for verification through a single inference step.
The verification process involves comparing the draft tokens with the LLM's output, where matched tokens are accepted. 
Notably, the LLM's inference result following the last accepted draft token serves as the input for the subsequent round of speculative decoding. 
While SLMs may not produce entirely accurate or comprehensive sequences, they demonstrate proficiency in generating simpler linguistic elements such as qualifiers, pronouns, and punctuation marks, which exhibit higher acceptance rates during verification.
When the draft sequence achieves sufficient accuracy, speculative decoding enables the LLM to produce multiple tokens in a single inference step, thereby substantially reducing the reliance on autoregressive inference and significantly enhancing overall inference speed.

The effectiveness of speculative decoding hinges on the alignment of output distributions between the SLM and the LLM.
To achieve this, some approaches \cite{leviathan2023fast} use models from the same family but of different sizes as the SLM and LLM (\eg, Vicuna-68M \vs Vicuna-7B).
However, those approaches impose constraints on model selection, limiting their flexibility. 
Therefore, alternative approaches \cite{li2024eagle} involve training lightweight, custom-designed models on small public datasets to serve as SLMs. 
Furthermore, recent advancements \cite{cai2024medusa} have introduced tree-based verification methods to optimize the number of accepted draft tokens at a single inference step.
Concretely, multiple candidate draft tokens at different positions are combined into a tree-like structure, where each branching sequence represents a potential path that may be accepted in the verification stage.

\textbf{U-shaped Inference.}
To safeguard the privacy of both input and output tokens on local devices, U-shaped inference has been proposed, which partitions the LLM into three submodels. 
The lightweight input and output submodels are deployed on end devices, while the computationally intensive middle submodel operates in the cloud. 
During inference, input tokens are first encoded into ``shallow'' hidden states by the local input submodel on each device.
Subsequently, these hidden states are transmitted to the cloud for further processing by the middle submodel.
Finally, the refined ``deep'' hidden states are transferred back to the device, where the output submodel decodes them into final predicted tokens.

U-shaped inference ensures raw input/output tokens remain confined to local devices, which enhances the privacy of raw data.
However, it introduces significant communication overhead, as each token generation requires bidirectional transmission of hidden states.
This overhead is exacerbated during \prefill phase, where long prompts produce large volumes of hidden states, causing severe communication delays.

\subsection{Motivation}\label{subsec_motivation}
To investigate the efficiencies of different inference frameworks, as well as the impact of long prompts on communication and in-cloud computation delays, we conduct preliminary experiments with the Vicuna model \cite{chiang2023vicuna} on the Specbench dataset \cite{xia-etal-2024-unlocking}.
In the experiments, a deep learning workstation equipped with 8 NVIDIA A6000 GPUs serves as the cloud server, and 3 NVIDIA Jetson AGX Orin devices are adopted as end devices.
For speculative decoding (SD), we separately deploy Vicuna-7B and Vicuna-68M on the server (for verification) and a device (for drafting).
Besides, for U-shaped inference, we deploy the first 2 decoder layers and the last head of Vicuna-7B on a device, and the remaining portion on the server.
As a comparison, we directly forward the request of a device to the server-side Vicuna-7B for inference as the cloud-based baseline.

\textbf{Delay of Different Inference Frameworks.}
There are two primary metrics for evaluating LLM inference speed, \ie, time-to-first-token (TTFT) and time-between-tokens (TBT). 
TTFT measures the delay of generating the first output token from the moment a request arrives in \prefill phase,
while TBT measures the time interval between the generation of consecutive output tokens during \decode phase.
Firstly, we illustrate the TTFT and the TBT (involving local computation delay, communication delay, and in-cloud computation delay) of different inference frameworks to process a request with the 128-token prompt in Figure \ref{fig: decode_time}, in which the proportion of computation delay and communication delay varies significantly among inference frameworks.
On one hand, SD makes full use of the local computational resources of devices to generate a multi-token draft sequence, which undergoes parallel verification by the in-cloud LLM, reducing the in-cloud computation delay and accelerating inference, compared to cloud-based inference.
On the other hand, since the size of hidden states is much larger than that of tokens, the communication delay for transmitting hidden states in U-shaped inference becomes substantial, especially in \prefill phase, leading to higher TTFT and TBT compared to both SD and cloud-based inference.

\textbf{Impact of Long Prompts on Communication.}
As shown in Figure \ref{fig: decode_time}, the communication delay in U-shaped inference accounts for more than 80\% of the TTFT in \prefill phase with the 128-token prompt.
We further explore the impact of varing prompt lengths (ranging from 128 to 2k tokens) on TTFT and communication delay in U-shaped inference.
By the results in Figure \ref{fig: Ushaped}, the communication delay increases almost linearly with the prompt length increasing and always occupies the majority of the TTFT.
Specifically, the communication delay of the 512-token prompt is 4$\times$ that of the 128-token prompt.
Besides, for the long prompts with 2k tokens, the TTFT is 3.57s, while the delays of local computation, in-cloud computation, and communication are 0.09s (2.5\%), 0.28s (7.8\%), and 3.20s (89.6\%), respectively.
Consequently, optimizing the handling of long prompts can yield substantial improvements in reducing communication delays for U-shaped inference and our system design.

\textbf{Impact of Long Prompts on In-Cloud Computation.}
In practice, in-cloud LLMs usually process requests from different phases (\prefill and \decode) simultaneously through batched inference.
According to our analysis in Section \ref{General Inference Process of LLMs}, 
requests with long prompts in \prefill phase can consume excessive computational resources and lead to increased in-cloud computation delay for other requests in \decode phase.
Therefore, we let the LLM process the batch consisting of one request with varying prompt lengths in \prefill phase and nine requests in \decode phase to explore the impact of prompt length on the in-cloud computation delay.
By Figure \ref{fig: batched_token_size}, as the prompt length increases, the in-cloud computation delay becomes higher and higher.
For instance, the in-cloud computation delay caused by 32-token prompt length is only 10.1\% higher than that caused by 1-token prompt length, while for prompt length more than 512 tokens, in-cloud computation delay almost increases linearly with the prompt length.
That is because batching \prefill phase with short prompts would improve GPU utilization with minimal delay increase, while long prompts would lead to saturation of computational resources and prolonged delays.

\textbf{Effect of Prompt Chunking.}
To reduce the in-cloud computation delay caused by long prompts, an intuitive solution is to divide long prompts into multiple chunked prompts (\ie, prompt chunking), which are processed separately across multiple inference steps.
To evaluate the effect of prompt chunking, we perform the consecutive 64-step batched inference by dividing a 2k-token prompt into multiple chunked prompts, in which the previous batches consist of one chunked prompt in \prefill phase as well as nine requests in \decode, and the remaining batches after processing the entire prompt only consist of ten requests in \decode.
The results are presented in Figure \ref{fig: chunk_size}, in which we record the reduction of total computation delay (\ie, the in-cloud computation delay for the consecutive 64-step batched inference) and the TTFT after processing all chunks in \prefill phase given different chunk sizes.
As evidenced by the results in Figure \ref{fig: chunk_size}, a decrease in chunk size correlates with an increase in the reduction of total computation delay and TTFT; however, the TTFT escalates at a faster rate. 
For example, with a chunk size of 32, the total delay is reduced by 82.3 ms, but the TTFT surges by approximately 6.6$\times$ compared to inference without prompt chunking.
In summary, while prompt chunking effectively alleviates in-cloud computation delays for devices in the \decode phase, it conversely results in an increased TTFT for devices in the \prefill phase. 
This dichotomy underscores the necessity of optimizing chunk size to strike an optimal balance for batched inference in the cloud

\begin{figure}[t]
\centering
\includegraphics[width=0.95\linewidth]{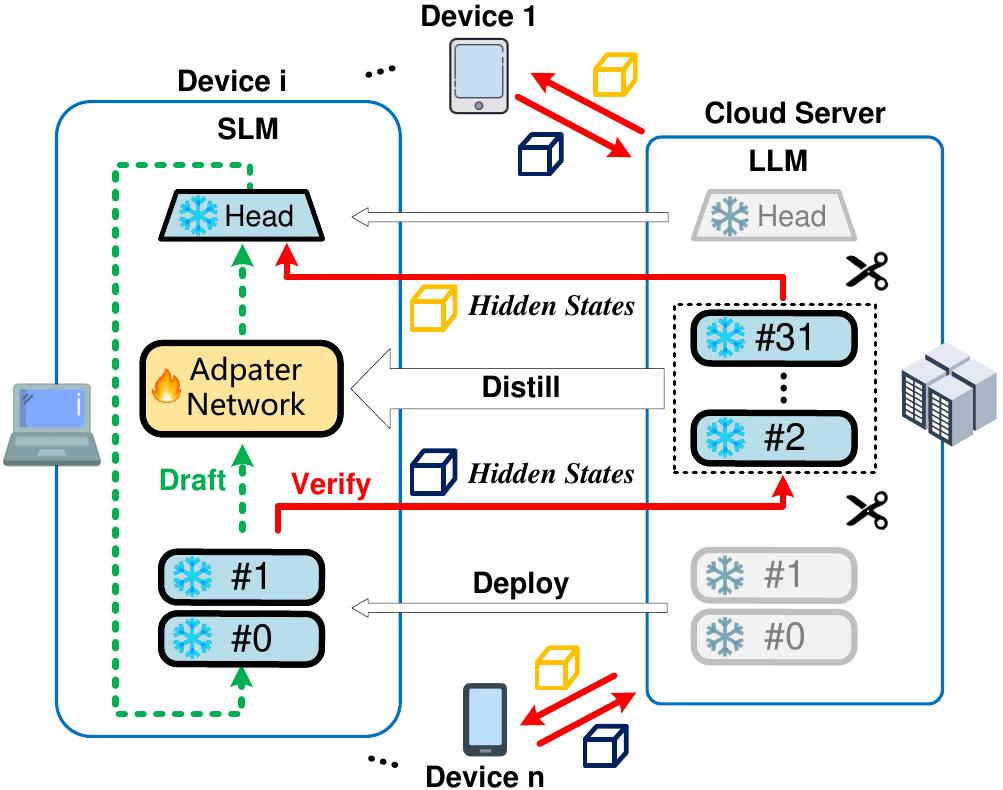}
\caption{Illustration of \method.}
\label{fig: SD}
\end{figure}

\subsection{Our Proposed Framework}
While U-shaped inference enhances the privacy protection of raw tokens, it inevitably introduces significant communication overhead due to the frequent transmission of hidden states. 
On the other hand, although speculative decoding reduces the device-cloud interaction frequency and accelerates inference speed, it presents inherent privacy vulnerabilities.
Considering the complementary strengths of U-shaped inference and speculative decoding, we propose to integrate U-shaped inference with speculative decoding and develop a new and novel device-cloud collaborative inference framework named \method.
Figure \ref{fig: SD} illustrates the \method framework, where an LLM with 32 decoder layers is exemplified.
In \method, the shallow layers (\eg, layers \#0 and \#1) and the head of LLM are deployed on end devices as input and output submodels to enhance privacy, while the middle submodel, encompassing other 30 decoder layers, resides in the cloud.
Besides, a lightweight adapter network distilled from the middle submodel is deployed on devices to form the draft model, which performs speculative decoding to accelerate inference and maintain promising accuracy.
As shown in Figure \ref{fig: SD}, the main inference process of \method involves local drafting (denoted in dashed green line) and device-cloud verification (denoted in solid red line), which looks like a ``top hat''.

According to the analysis in Sec. \ref{subsec_motivation}, chunking the long prompts into multiple short prompts can alleviate in-cloud computation delays for devices in \decode phase.
Upon the multiple short prompts, \method can maximize the overlap between device-side transmission of chunked hidden states and in-cloud computation of previously adjacent chunked hidden states to reduce the TTFT for devices in \prefill phase (more details in Sec. \ref{Prompt Chunking}).
Besides, considering the characteristic of speculative decoding, \method allows devices to continue local drafting (\ie, parallel drafting) while awaiting verification results from the LLM in \decode phase so as to further utilize the computational resources of devices and improve inference speed (more details in Sec. \ref{Parallel Drafting}).

\section{System Design}\label{sec:alg}
\subsection{System Overview}
\begin{figure}[t]
\centering
\includegraphics[width=1.0\linewidth]{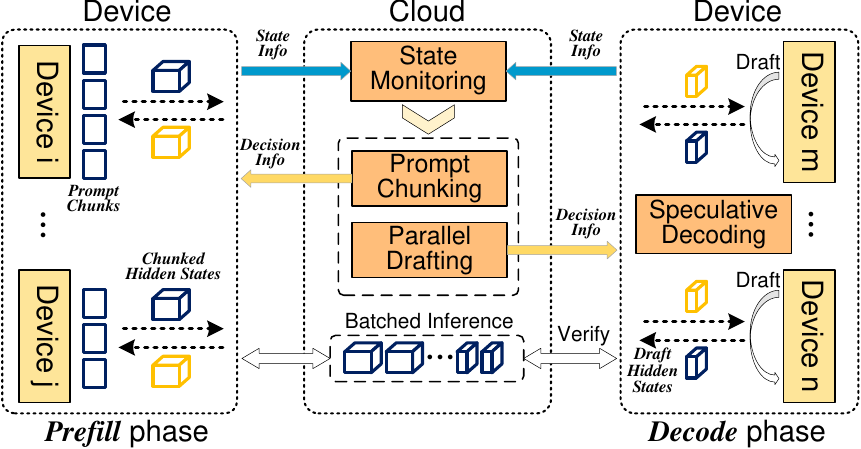}
\vspace{-0.5cm}
\caption{System overview of \method.}
\label{fig: overview}
\end{figure}

The system design of \method is illustrated in Figure \ref{fig: overview}. 
\method consists of four key modules, \ie, \textit{state monitoring}, \textit{prompt chunking}, \textit{speculative decoding}, and \textit{parallel drafting}.
Through batched inference, the cloud processes requests from multiple devices in \prefill phase and \decode phase  simultaneously.
At regular intervals, the state monitoring module gathers critical data on the cloud's workload and the state information of all devices, including their computing and communication capabilities.
When a request arrives at a device, the prompt chunking module determines an optimal chunk size tailored to the device's capabilities to complete \prefill phase, in which the device divides the prompt into specified chunks and processes them in a pipelined fashion.
Following the \prefill phase, the speculative decoding module orchestrates an iterative process where the device drafts potential continuations and verifies them against the cloud's predictions in \decode phase.
Concurrently, the parallel drafting module facilitates the device in generating a draft sequence for the subsequent round of speculative decoding during the verification stage.

\subsection{State Monitoring Module}

To determine the appropriate chunk size, it is necessary to continuously collect multiple rounds of state information of the cloud (\eg, workload) and all devices (\eg, computing and communication capabilities).

Concretely, the cloud serves requests from different devices at different moments, resulting in a dynamic workload.
We denote the batched token size (\ie, the number of processed tokens in a batch) and the in-cloud computation delay for the batch in round $t$ as $\mu^{t}$ and $\eta^{t}$, respectively.
As proxy metrics, we adopt $\mu^{t}$ and $\eta^{t}$, which can be recorded by the cloud directly, to indicate the workload of the cloud in round $t$.
Besides, the cloud maintains a predictive function $g^{t}(\cdot)$ 
for predicting the in-cloud computation delays associated with different chunk sizes.
During the information collection of round $t$, the cloud collects the latest $\hat{\mu}^{t}$ and $\hat{\eta}^{t}$, and updates the predictive function $g^{t}(\cdot)$. 
Moreover, to improve the robustness of the estimation, we introduce moving averages with historical workload \cite{leroy2019federated}.
Accordingly, the cloud estimates $\mu^{t}$ and $g^{t}(\cdot)$ in round $t$ by calculating the moving average with $\alpha \in [0,1]$ (\eg, 0.8 in our experiments) as:
\begin{equation}\label{eq:mu}
\vspace{-0.1cm}
\begin{aligned}
\mu^{t} = \alpha \cdot \mu^{t-1} + (1-\alpha) \cdot \hat{\mu}^{t}
\end{aligned}
\vspace{-0.05cm}
\end{equation}
\begin{equation}\label{eq:g}
\begin{aligned}
g^{t}(\mu^{t}) = \alpha \cdot g^{t-1}(\mu^{t}) + (1-\alpha) \cdot \hat{\eta}^{t}
\end{aligned}
\end{equation}
With this predictive function, the cloud can predict the computation delay based on batched token size, enabling more efficient chunk size optimization. 

Additionally, the collection of time-varying computing and communication capabilities of devices is necessary for chunk size optimization.
We use $\gamma^{t}_{i}$, $\beta^{t}_{i, up}$, and $\beta^{t}_{i, down}$ to denote the drafting delay, uploading, and downloading bandwidths of device $i$ in round $t$, respectively, which can be recorded and computed by devices directly.
Besides, we also compute their moving average based on their historical information.
Since the size of state information is much smaller than that of hidden states, it is reasonable to ignore the computation/communication overhead for state monitoring \cite{lyu2019optimal}.

\subsection{Prompt Chunking Module}\label{Prompt Chunking}

\begin{figure}[t]
\centering
\includegraphics[width=0.98\linewidth]{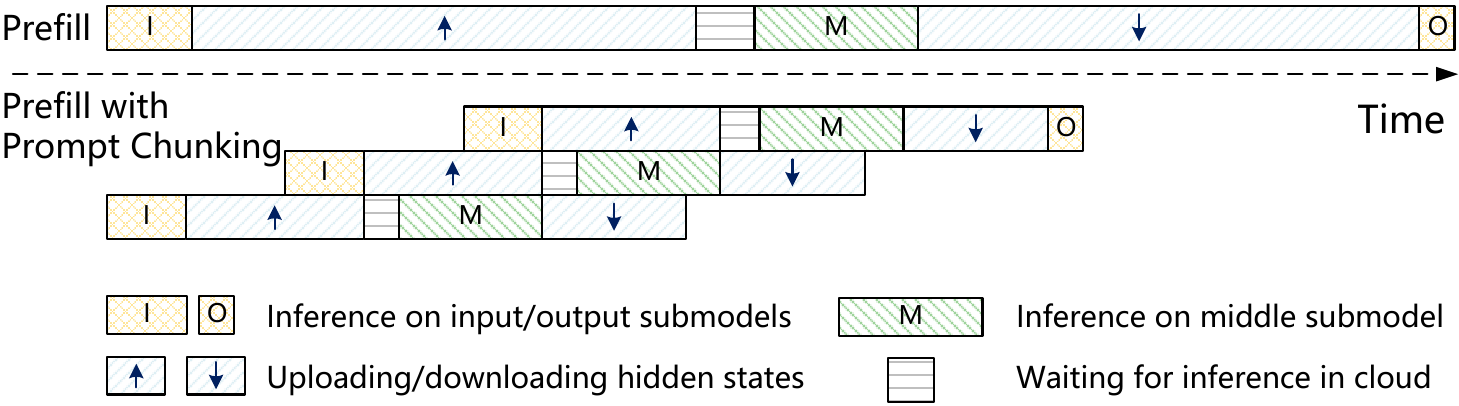}
\caption{Illustration of Prompt Chunking.}
\label{fig_chunk_prompt}
\end{figure}

Upon the collected state information, the cloud determines diverse and appropriate chunk sizes for the devices to complete prefilling with prompt chunking.
Figure \ref{fig_chunk_prompt} illustrates the differences in processing the prompts of a device with/without the prompt chunking mechanism in \prefill phase.
Concretely, this mechanism enables the device to divide its long prompt into multiple shorter chunks with a certain chunk size, and process these chunks in a pipelined manner.
By parallelizing the computation and transmission of hidden states, the uploading of chunked hidden states can overlap with other operations like in-cloud computation and downloading, thereby reducing the delay in \prefill phase. 
Moreover, processing chunked hidden states in the cloud demands fewer computational resources, which incurs less in-cloud computation delay for other devices in \decode phase.

Given a long prompt, the chunk size critically influences both communication and in-cloud computation delays, and the total number of chunks. 
As indicated in Sec. \ref{subsec_motivation}, although communication delay exhibits an approximately linear relationship with chunk size, the in-cloud computation delay demonstrates limited reduction with decreasing chunk size, particularly at smaller scales. 
This phenomenon can result in accumulated computation delays across all chunks that exceed the communication delay within the pipeline framework.
Conversely, larger chunk sizes may introduce substantial per-chunk delays, thereby reducing parallelization efficiency. 
Therefore, optimal chunk size selection should balance granularity to minimize the in-cloud computation delay for each chunk, while maintaining sufficient communication delay to overlap with preceding chunk processing.

Furthermore, each incoming chunk experiences a delay in waiting for the current batch to be processed in the cloud, making the total in-cloud delay the sum of waiting and in-cloud computation delays.
The maximum waiting delay approximates the average in-cloud computation delay $g^{t}(\mu^{t})$, while the computation delay for a chunk of size $X_i$ from device $i$ is estimated as $g^{t}(\mu^{t}+X_i)$.
In practice, the cloud usually utilizes multiple GPUs for pipeline-parallel inference, where computation delay per GPU is inversely proportional to the number of GPUs (denoted as pipeline length). 
This architecture eliminates the need to wait for the previous inference to be finished across the entire model in the cloud.
Accordingly, the optimal chunk size $X_i$ is calculated with pipeline length $P$ by the following formula:
\vspace{-0.1cm}
\begin{equation}\label{eq:x}
\begin{aligned}
\frac{X_i \cdot A}{\beta^{t}_{i,up}} = \frac{g^{t}(\mu^{t}) + g^{t}(\mu^{t}+X_i)}{P}
\end{aligned}
\vspace{-0.1cm}
\end{equation}
where $A$ is the size of the hidden state per token.
We will further explore the impact of pipeline length scale on inference delay of \method in Sec. \ref{Effect of Pipeline Length Scale}.

\subsection{Speculative Decoding Module}
We first introduce the construction and training of the lightweight adapter network.
Generally, given an LLM $w^{n}_{L}$, which consists of an output head and $n$ decoder layers.
The lightweight adapter network $\Lambda$ is constructed and combined with the LLM's output head $H_{L}$ and first $m$ decoder layers $w^{m}_{L}$ ($m$ less than 10\% of $n$) to form a draft model $w_{S} =  H_{L} \circ \Lambda \circ w^{m}_{L}$.
$\Lambda$ has the same structure as the self-attention module of the decoder layer, as the self-attention module has fewer parameters and computation delay than the feed-forward network.
To enable efficient drafting, $\Lambda$ is trained by knowledge distillation to learn the middle submodel of $w^{n}_{L}$.
Following \cite{li2024eagle}, we use Smooth L1 loss ($L_{sl}$) and  Cross-Entropy loss ($L_{ce}$) as the training loss function for knowledge distillation:
\begin{equation}\label{eq:loss}
\begin{aligned}
\vspace{-0.05cm}
Loss = L_{sl}(f^{L}_{i+1}, f^{S}_{i+1}) + w_{ce} L_{ce}(H_{L}(f^{L}_{i+1}), H_{L}(f^{S}_{i+1}))
\vspace{-0.05cm}
\end{aligned}
\end{equation}
where $f^{L}_{i+1}$  and $f^{S}_{i+1}$ denote the hidden states of the new token $t_{i+1}$ in $w^{n}_{L}$ and $w_{S}$, respectively, before they are fed into the output head. 
The weight $w_{ce}$ controls the contribution of Cross-Entropy loss and is typically set to 0.1.

The draft model $w_{S
}$ and LLM $w^{n}_{L}$ collaborate in performing multiple rounds of speculative decoding.
Specifically, each round of speculative decoding begins with the drafting stage, where $w_{S}$ on the device performs multiple autoregressive inferences to generate a draft sequence. 
During the subsequent verification stage, the draft sequence is first processed by the shallow layers $w^{m}_{L}$ to produce shallow hidden states, which are then transmitted to the cloud for further inference by the middle submodel of $w^{n}_{L}$ to produce deep hidden states.
The deep hidden states are then sent back to the device, where the final inference result is generated by the LLM's head $H_{L}$.
The draft tokens with the same inference result of LLM will be accepted, and a new token will be generated from the last accepted draft token.
Subsequently, this new token is used as input for the next round of speculative decoding.

During the drafting stage, the efficiency of speculative decoding is greatly influenced by the drafting sequence length $n$, which reflects the number of drafting steps.
Concretely, if the draft sequence length $n$ is too short, some simpler tokens that would have been accepted may be missed.
Conversely, a long draft sequence length can lead to the squandering of computational resources on challenging tokens, increasing drafting overhead.
To optimize the drafting process, we implement a drafting threshold $\eta \in [0,1]$ (\eg, 0.6 in our experiments) to adaptively stop the draft when the softmax probability of the draft token falls below the threshold :
\begin{equation}\label{eq:draftlen}
\vspace{-0.1cm}
\begin{aligned}
\mathop{max}\limits_{n} w_{S}(t_{i+n}) < \eta
\end{aligned}
\end{equation}
where $w_{S}(t_{i+n})$ denoted the softmax probability of the $n$-th token generated by $w_{S}$ during the drafting stage.

\subsection{Parallel Drafting Module}\label{Parallel Drafting}

\begin{figure}[!t]
\centering
\includegraphics[width=0.85\linewidth]{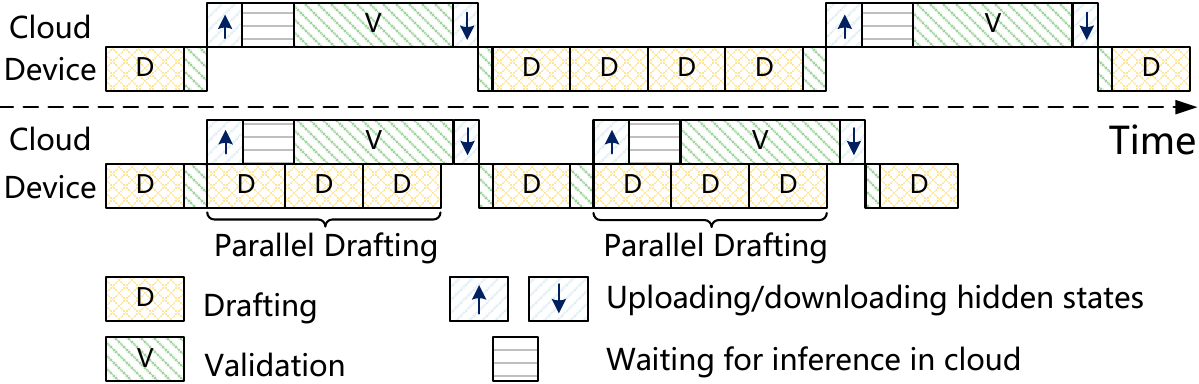}
\caption{Illustration of Parallel Drafting.}
\label{fig_para_decode}
\end{figure}

Due to the limited computing capabilities of devices, the local computation for drafting incurs non-negligible delays, reducing the efficiency of speculative decoding.
Therefore, we introduce the parallel drafting mechanism to parallel the drafting and verification stages, as shown in Figure \ref{fig_para_decode}.
This mechanism enables devices to generate the draft sequence for the next round of speculative decoding in advance, fully utilizing idle computational resources of devices during the verification stage.

Specifically, the device takes the top-k tokens generated from the last inference step in the drafting stage as candidate tokens.
During the verification stage, these candidate tokens serve as the initial input to generate candidate draft sequences in parallel.
According to Eq. (\ref{eq:draftlen}), the softmax probability of the last draft token falls below the drafting threshold, resulting in a relatively low probability of being accepted.
Therefore, the last draft token may be corrected into a new token in the verification stage.
If this new token matches one of the top-k candidate tokens, the corresponding candidate draft sequence will be used for the subsequent round of speculative decoding.

Based on the collected state information, the cloud selects an appropriate number of inference steps to guide the devices to generate candidate draft sequences during the verification stage.
To fully utilize the idle computational resources of devices without interference with the verification stage, the generation process should be completed before receiving the inference result from the cloud.
Therefore, we consider the case where the in-cloud delay of the verification stage is minimized (\ie, no wait time in the cloud).
We use $\lambda_i$ to represent the number of inference steps for parallel drafting in device $i$, which is calculated as follows:
\begin{equation}\label{eq:x}
\begin{aligned}
\lambda_i = \lfloor \frac{\frac{\mu_i \cdot A}{\beta^{t}_{i,up}} + g^{t}(\mu^{t})  + \frac{\mu_i \cdot A}{\beta^{t}_{i,down}}}{\gamma^{t}_{i}} \rfloor
\end{aligned}
\end{equation}
where $\mu_i$ is the draft sequence length of device $i$ in the current round of speculative decoding.


\section{Experiments and Evaluation}\label{sec:results}

\subsection{Experimental Settings}

\textbf{System Deployment.} We conduct extensive experiments to evaluate the performance of \method on an end-to-end hardware prototype system.
Specifically, we employ a deep learning workstation as the cloud server, which is equipped with an Intel(R) Xeon(R) Platinum 8358P CPU, 8 NVIDIA GeForce RTX A6000 GPUs, and 512 GB RAM.
In addition, we specify 30 NVIDIA Jetson kits, including 20 Jetson AGX Xavier and 10 Jetson AGX Orin as devices to construct a heterogeneous system. 
The detailed technical specifications of Jetson AGX Xavier and Orin are listed in Table \ref{table:jetson}.

\begin{table}[t]
\caption{Device Technical Specifications.}
\label{table:jetson}
\centering
\begin{tabular}{lcc}
    \hline
     & \textbf{AI Performance} & \textbf{GPU Type}  \\ 
     \hline
    AGX Xavier & 32 TOPS & 512-core Volta  \\  
    AGX  Orin & 200 TOPS & 1792-core Ampere  \\  
    \hline \hline 
    & \textbf{CPU Type} & \textbf{ROM} \\  
    \hline
    AGX Xavier  & 8-core Carmel ARM 8 & 32 GB LPDDR4x \\  
    AGX  Orin  & 8-core Cortex ARM 8 & 32 GB LPDDR5 \\  
    \hline
\end{tabular}
\end{table}

\begin{table}[!t]
\caption{Prompt Token Length and Models.}
\label{table:datasets}
\centering
\begin{tabular}{ccccc}
\hline
    \textbf{Dataset}  & \textbf{Avg} & \textbf{P90} & \textbf{Std} & \textbf{Model} \\ 
    \hline
    SpecBench  & 351.2 & 891.0 & 397.3 & Vicuna-7B \\  
    CNN/DM  & 1036.6 & 1772.0 & 511.8 & Vicuna-13B \\  
    \hline 
\end{tabular}
\end{table}

In the experiments, we build the software platform based on Docker Swarm \cite{merkel2014docker, naik2016building} and the PyTorch deep learning library \cite{paszke2019pytorch}. 
The Docker Swarm, a distributed software development kit, facilitates the construction of a distributed system and enables the monitoring of each device’s operational status.
The PyTorch library facilitates the implementation of model inference across devices. 
Additionally, MPI (Message Passing Interface) \cite{gabriel2004open} which includes a collection of sending and receiving functions, is implemented to streamline communication between devices and the server.
Furthermore, we incorporate FlashAttention \cite{dao2023flashattention} and FlashInfer \cite{ye2024accelerating} kernels to support page attention and continuous batching, and we use NCCL \cite{NCCL} for pipeline communication between GPUs.

\textbf{Setting of System Heterogeneity.}
To enable the devices with heterogeneous computing and communication capabilities, we present the following experimental settings.

1) \textbf{\textit{For Computation.}}
All the Jetson devices can be configured to work with different modes, specifying the number of working CPUs and the frequency of CPU/GPU, so that they can work with different computing capabilities. 
For instance, the AGX Orin with the highest performance mode (\ie, mode 0) can achieve inference by 10$\times$ faster than the AGX Xavier with the lowest performance mode (\ie, mode 1).
To simulate the time-varying resources of devices, we randomly change the modes of devices after every 5 requests generated.

2) \textbf{\textit{For Communication.}}
All devices are connected to the server via WiFi routers.
We group devices into three groups, each containing 10 devices. 
These groups are then placed at different locations, \ie, 2m, 8m, and 14m away from the WiFi routers. 
Due to random channel noise and competition among devices, the bandwidth between the server and devices dynamically varies. 
Devices' bandwidth ranges from 5MB/s to 10MB/s for uplinks and from 10MB/s to 15MB/s for downlinks, as measured by iperf3 \cite{tirumala1999iperf}.

\begin{figure}[t]
	\centering
	\subfigure[TTFT]
	{
		\includegraphics[width=0.46\linewidth]{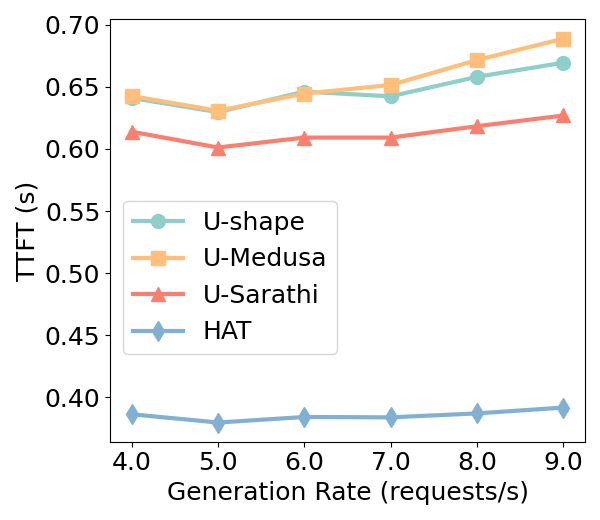}
		\label{fig: SpecBench_rate_TTFT}
	}
	\subfigure[TBT]
	{
		\includegraphics[width=0.46\linewidth]{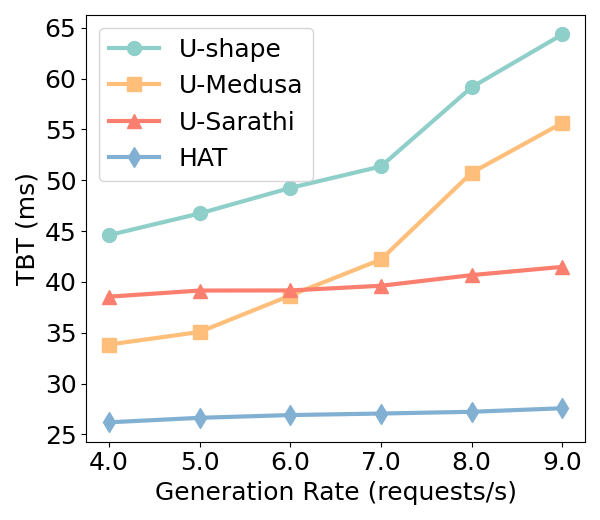}
		\label{fig: SpecBench_rate_TBT}
	}
        \vspace{-0.3cm}
	\caption{Delay with Different Request Generation Rates on SpecBench.}
	\vspace{-0.1cm}
	\label{lab: rate_specbench}
\end{figure}

\textbf{Datasets and Models.}
We evaluate \method's performance on two datasets with Vicuna model fine-tuned from LLama.

1) \textbf{\textit{SpecBench}} \cite{xia-etal-2024-unlocking} is a versatile dataset encompassing a range of tasks such as translation, summarization, mathematical reasoning, \etc. 
The tasks in SpecBench exhibit significant variations in prompt length.
For instance, the prompts for summarisation tasks have an average length of 877.3, while the average prompt length for the translation tasks is only 82.5.
For this dataset, we adopt a famous Vicuna-7B LLM, which includes 32 decoder layers with 32 attention heads, and its hidden size is 4096.

2) \textbf{\textit{CNN/DM}} \cite{see-etal-2017-get} focuses on text summarization and comprises 11490 news articles, each associated with long prompts.
Concretely, the tasks of CNN/DM have prompts with an average length of 1036.6.
The specific prompt token lengths for each dataset are listed in Table \ref{table:datasets}.
For this dataset, we utilize a more robust Vicuna-13B LLM, which includes 40 decoder layers with 40 attention heads, and its hidden size is 5120.

\textbf{Baselines.} We measure the effectiveness of \method through a comparison with three baselines.

1) \textbf{\textit{U-shape}} \cite{han2023federated, sa2024ensuring}\textbf{.}  It is a famous split inference approach that deploys LLM's head (output submodel) and shallow layers (input submodel) to devices to enhance privacy, while the computationally intensive middle submodel resides in the server.
Note that given our primary focus on privacy-preserving inference, all baselines are adapted within the U-shaped framework. 
We specifically emphasize the comparative analysis of inference efficiency across various modified frameworks.

2) \textbf{\textit{U-Medusa.}} We integrate Medusa \cite{cai2024medusa}, an efficient speculative decoding approach, within the U-shaped inference framework as U-Medusa.
Medusa constructs and trains 4 Medusa heads to predict the next 4 tokens per inference step.
Therefore, U-Medusa deploys 4 Medusa heads with LLM's input and output submodels to devices to enhance privacy.
Additionally, it employs a tree verification method that aggregates candidate tokens from Medusa heads to extend the number of accepted draft tokens.

3) \textbf{\textit{U-Sarathi.}} We integrate Sarathi-Serve \cite{agrawal2024taming}, a state-of-the-art LLM serving system, within the U-shaped inference framework as U-Sarathi.
The server in U-Sarathi chunks prompts to fully utilize its computational resources, minimizing the bubbles in pipeline-parallel inference and reducing the computation delay for devices in \decode phase.

\textbf{Metrics.}
There are two primary delay metrics for LLM inference performance: time-to-first-token (TTFT) and time-between-tokens (TBT).
TTFT measures the delay of generating the first output token from the moment a request arrives in \prefill phase, which reflects the initial responsiveness of the system.
Besides, TBT measures the interval between the generation of consecutive output tokens during \decode phase, which affects the overall perceived fluidity of the response.
Moreover, LLM inference services must meet stringent delay requirements to maintain the quality of service.
Accordingly, we establish two service level agreements (SLAs) for \prefill and \decode to assess the SLA compliance rate, \ie, the percentage of requests that satisfy these SLA targets.


\textbf{Experimental Parameters.}
We deploy the first 2 shallow layers of the Vicuna-7B, and the first 3 shallow layers of the Vicuna-13B, along with their respective heads, to devices. 
In addition, considering the trade-off between the communication delay of hidden states and accept length, we choose tree verification of size 8 for U-Medusa to enhance the efficiency of speculative decoding. 
For U-Sarathi, we choose chunk sizes of 128 and 256 for SpecBench and CNN/DM datasets, respectively.
Moreover, both the U-Medusa heads and \method's constructed network $\Lambda$ are trained on the ShareGPT dataset \cite{ShareGPT}.
Furthermore, the maximum generation lengths for all datasets are set to 128 tokens.

\subsection{Overall Performance}

\begin{figure}[!t]
\centering
\subfigure[TTFT]
{
    \includegraphics[width=0.46\linewidth]{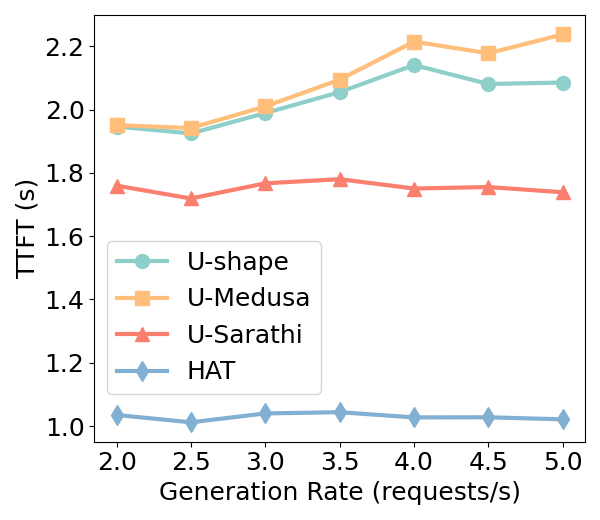}
    \label{fig: CNNDM_rate_TTFT}
}
\subfigure[TBT]
{
    \includegraphics[width=0.46\linewidth]{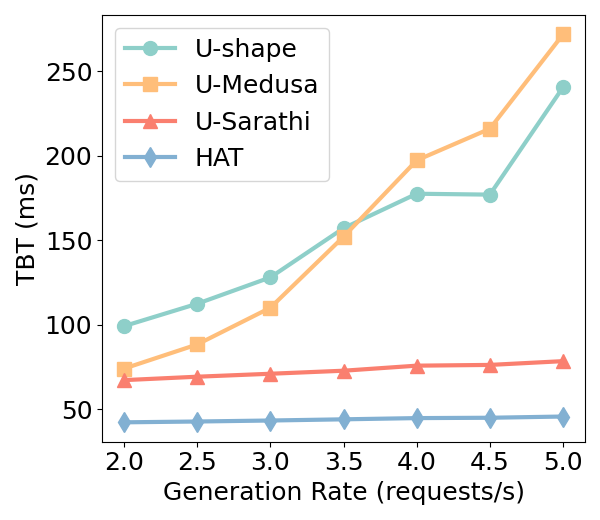}
    \label{fig: CNNDM_rate_TBT}
}
\vspace{-0.3cm}
\caption{Delay with Different Request Generation Rates on CNN/DM.}
\vspace{-0.1cm}
\label{lab: rate_CNNDM}
\end{figure}

Firstly, we conduct a set of experiments on two datasets with different request generation rates (\ie, the number of requests per second) to evaluate the performance of \method and baselines.
In the experiments, devices generate requests following a Poisson process, and the server employs 4 GPUs for pipeline-parallel inference (\ie, the server's pipeline length is 4).
As illustrated in Figures \ref{lab: rate_specbench}-\ref{lab: rate_CNNDM}, \method consistently achieves the lowest TTFT and TBT among all approaches. 
For instance, Figure \ref{fig: SpecBench_rate_TTFT} shows that at a generation rate of 6 requests/s on SpecBench, \method achieves the TTFT of 384.2ms, significantly lower than the TTFT of U-Sarathi (609.2ms), U-Medusa (644.6ms), and U-shape (646.2ms).
Moreover, by Figure \ref{fig: SpecBench_rate_TBT}, when the generation rate is 4 requests/s, \method reduces the TBT by 31.9\%, 22.5\% and 41.3\%, compared to U-Sarathi, U-Medusa and U-shape, respectively.
Besides, as the generation rate increases from 4 to 9 requests/s, the TBT for \method and U-Sarathi increase only 5.3\% and 7.8\%, respectively, while U-Medusa and U-shape experience more substantial increases of 64.5\% and 44\%.
\method and U-Sarathi can mitigate the interference between multiple requests through chunking prompts, resulting in more stable TBT.
However, \method further reduces the TTFT and the TBT by parallel computation and communication along with its efficient speculative decoding.
In contrast, the tree verification method employed by U-Medusa, which aggregates more candidate draft tokens to extend the accept length, tends to overload the server.
Specifically, by Figure \ref{fig: CNNDM_rate_TTFT}, at a generation rate of 4 requests/s on CNN/DM, \method achieves the TTFT of 1027.4ms, while the TTFT of U-Sarathi, U-Medusa and U-shape is 1750.6ms, 2214.8ms and 2140.8ms, respectively. 
Besides, by Figure \ref{fig: CNNDM_rate_TBT}, with the same generation rate, \method can reduce the TBT by 40.8\%, 77.2\%, and 74.7\%, compared to the U-Sarathi, U-Medusa, and U-shape, respectively.
These results demonstrate that \method is effective under varying request loads.

\begin{figure}[!t]
\centering
\subfigure[SpecBench]
{
    \includegraphics[width=0.46\linewidth]{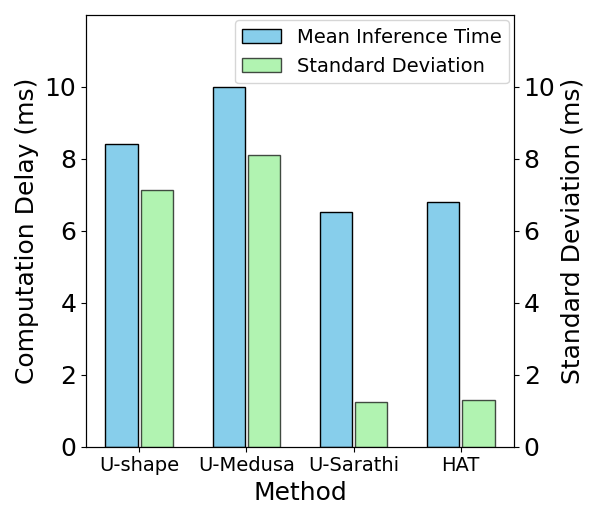}
    \label{fig: SpecBench_infer}
}
\subfigure[CNN/DM]
{
    \includegraphics[width=0.46\linewidth]{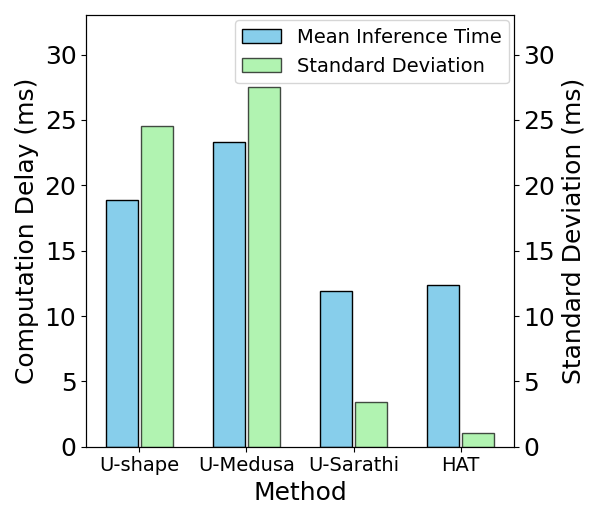}
    \label{fig: CNNDM_infer}
}
\vspace{-0.3cm}
\caption{Computation Delay with Standard Deviation.}
\vspace{-0.1cm}
\label{lab: infer}
\end{figure}


Secondly, to further demonstrate the efficiency of \method, we present the in-cloud computation delay of each GPU in Figure \ref{lab: infer}. 
Both \method and U-Sarathi achieve stable and low computation delay by chunking long prompts.
For instance, Figure \ref{fig: SpecBench_infer} shows that the average computation delay of \method  on SpecBench is 6.8ms, while U-Sarathi, U-Medusa, and U-shape incur that of 6.5ms, 10.0ms, and 8.4ms, respectively.
Besides, \method maintains a standard deviation in computation delay of 1.3ms, significantly lower than the 8.1ms and 7.1ms observed for U-Medusa and U-shape, respectively, and slightly above the 1.2ms for U-Sarathi.
U-Medusa and U-shape can not mitigate the impact of long prompts on the computation delay, leading to more volatile delays.
Specifically, by Figure \ref{fig: CNNDM_infer}, the average computation delay for \method on CNN/DM is 12.4ms with a standard deviation of 1.1ms, while U-Sarathi, U-Medusa, and U-shape exhibit the average computation delay of 11.9ms, 23.3ms, and 18.9ms, with corresponding standard deviations of 3.4ms, 27.5ms, and 24.6ms, respectively.
These results illustrate that \method is effective in ensuring stable inference performance.


\begin{figure}[!t]
\centering
\subfigure[Prefill SLA]
{
    \includegraphics[width=0.46\linewidth]{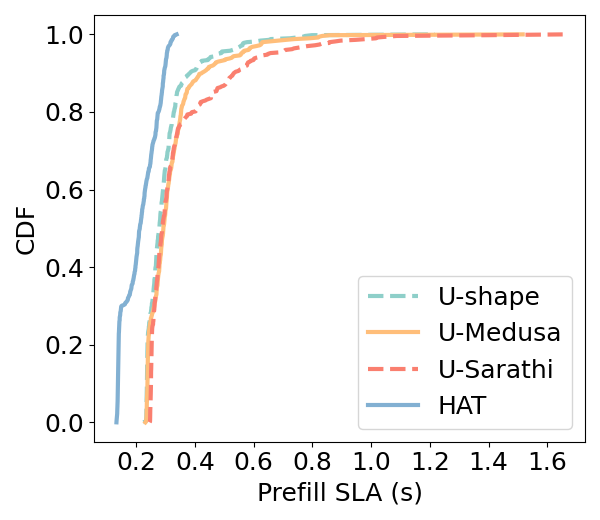}
    \label{fig: SpecBench_cdf_TTFT}
}
\subfigure[Decode SLA]
{
    \includegraphics[width=0.46\linewidth]{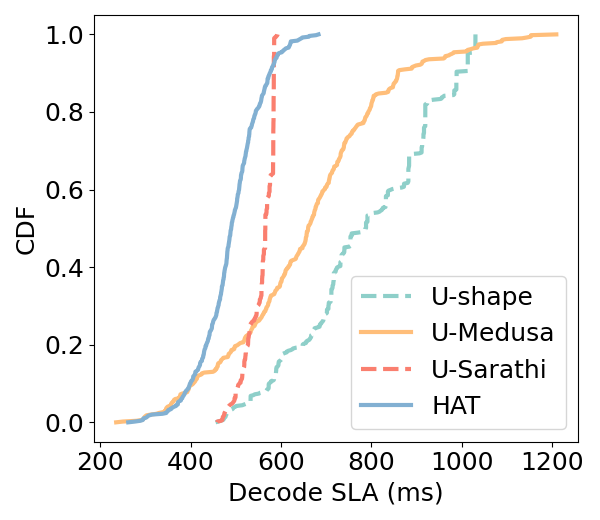}
    \label{fig: SpecBench_cdf_TBT}
}
\vspace{-0.3cm}
\caption{Compliance Rate with Different SLA on SpecBench.}
\vspace{-0.1cm}
\label{lab: cdf_specbench}
\vspace{-0.3cm}
\end{figure}

\begin{figure}[!t]
\centering
\subfigure[Prefill SLA]
{
    \includegraphics[width=0.46\linewidth]{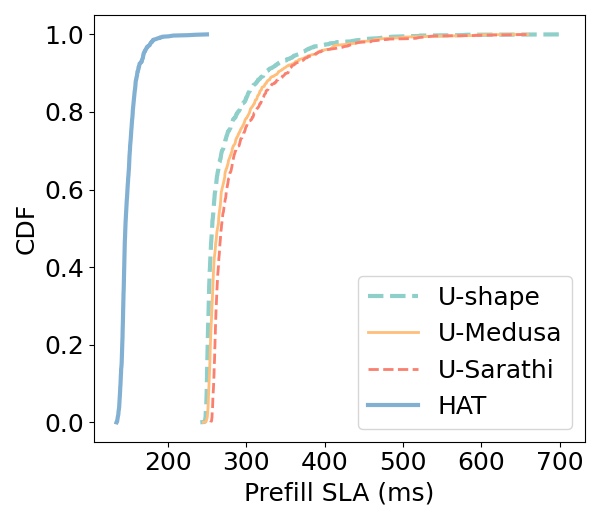}
    \label{fig: CNNDM_cdf_TTFT}
}
\subfigure[Decode SLA]
{
    \includegraphics[width=0.46\linewidth]{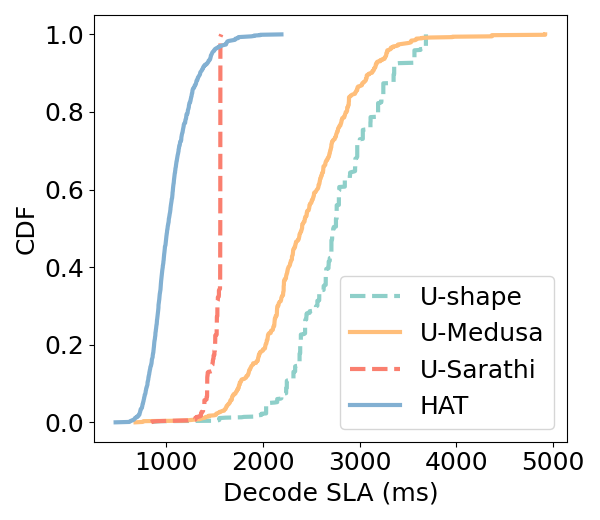}
    \label{fig: CNNDM_cdf_TBT}
}
\vspace{-0.3cm}
\caption{Compliance Rate with Different SLA on CNN/DM.}
\vspace{-0.1cm}
\label{lab: cdf_CNNDM}
\end{figure}


Thirdly, to illustrate the advantage of \method in maintaining the quality of service, we evaluate the SLA compliance rate, \ie, the proportion of requests that satisfy the SLAs for \prefill and \decode.
Concretely, the \prefill SLA is defined as the delay for processing per 128 prompt tokens, and the \decode SLA is defined as the delay for generating per 10 tokens in \decode phase. 
With a specific request generation rate and the server's pipeline length of 1, we monitor the delay of each token generation to ascertain SLA compliance.
The CDF plots displaying the SLA compliance rate with different SLA constraints for the four approaches are shown in Figures \ref{lab: cdf_specbench}-\ref{lab: cdf_CNNDM}. 
By the results, \method achieves the optimal compliance rate across most SLA cases.
For instance, as illustrated in Figure \ref{fig: SpecBench_cdf_TTFT}, when the \prefill SLA on SpecBench is 350ms, \method reaches 100\% of compliance rate, while the compliance rate of U-Sarathi, U-Medusa, and U-shape is 76.8\%, 78.8\% and 86.6\%, respectively. 
Besides, by Figure \ref{fig: SpecBench_cdf_TBT}, 50\% of requests in \method meet the \decode SLA of 489.3ms, while 50\% of requests in U-Sarathi, U-Medusa, and U-shape can satisfy more relaxed \decode SLAs of 565.0ms, 659.8ms, and 786.1ms, respectively.
\method significantly reduces the TTFT through the prompt chunking mechanism, and its advantage is more obvious for requests with long prompts.
Sarathi maintains a stable decoding rate through chunking, but it does not parallel transmission and computation, resulting in a longer delay in \prefill phase. 
Specifically, as illustrated in Figure \ref{fig: CNNDM_cdf_TTFT}, when the \prefill SLA on CNN/DM is 300ms, \method maintains a 100\% compliance rate, while U-Sarathi, U-Medusa, and U-shape exhibit lower rates of  64.7\%, 68.9\%, and 76.3\%, respectively. 
Besides, \method can outperform other baselines in most \decode SLA cases through efficient speculative decode, by Figure \ref{fig: CNNDM_cdf_TBT}, 90\% of requests in \method meet a \decode SLA of 1352.7ms, while 90\% of requests in U-Sarathi, U-Medusa, and U-shape satisfy more relaxed \decode SLAs of 1561.5ms, 3109.6ms, and 3357.8ms, respectively.

\subsection{SD Performance Evaluation}

\begin{table}[t]
\caption{SD Performance Evaluation.}
\label{table:SD eval}
\centering
\scalebox{0.95}{
\begin{tabular}{c|cccc}
    \hline
    \textbf{Dataset}  & \textbf{Method} & \textbf{Params} & \textbf{Accept} & \textbf{Speedup} \\ 
    \hline
    \multirow{3}{*}{\parbox{1.8cm}{SpecBench\\(Vicuna-7B)}} & U-shape  & N/A & 1.00    & 1.00$\times$ \\  
                                                          & U-Medusa & 591M & 1.89 & 1.41$\times$ \\  
                                                          & \method    & 67M & 2.06 & 1.65$\times$ \\  
    \hline 
    \multirow{3}{*}{\parbox{1.9cm}{CNN/DM\\(Vicuna-13B)}}   & U-shape  & N/A & 1.00    & 1.00$\times$ \\ 
                                                          & U-Medusa & 760M & 1.75 & 1.45$\times$ \\ 
                                                          & \method    & 105M & 1.98 & 1.60$\times$ \\  
    \hline 
\end{tabular}}
\end{table}

In this subsection, we evaluate the effectiveness of speculative decoding of \method and U-Medusa across two datasets.
We use the U-shape as a baseline to illustrate the speedup in \decode phase, and we also compare the accept length (\ie, number of accepted draft tokens) and the number of trained parameters.
In this experiment, we use only one device collaborating with the server to eliminate wait delay and prevent interference from other devices.
As shown in Table \ref{table:SD eval}, \method consistently outperformed U-Medusa.
For example, the \method on SpecBench has an accept length of 2.06, and achieves a decoding speedup of 1.65$\times$ compared to U-shape, while the accept length and speedup of U-Medusa are only 1.89 and 1.41$\times$, respectively. 
Besides, the \method network has more efficient trained parameters, requiring only 67M parameters for training compared to U-Medusa's 591M parameters.
The tree verification method of U-Medusa leads to a higher communication delay compared to \method's threshold-based method.
Concretely, for Vicuna-13B on CNN/DM, \method achieves 1.60$\times$ speedup compared to U-shape, and its accept length is 1.98 with only 105M trained parameters, while U-Medusa needs to train 760M parameters and its accept length is only 1.75 and achieves 1.45$\times$ speedups.
These results further demonstrate the advantage of \method with efficient speculative decoding.


\subsection{Effect of Key Strategies}


\begin{table}[t]
\caption{Effect of Key Strategies.}
\label{table:Ablation Study}
\centering
\scalebox{0.95}{
\begin{tabular}{l|ccccc}
\hline
\textbf{Dataset}  & \textbf{SD} & \textbf{PC} & \textbf{PD} & \textbf{TTFT(ms)} & \textbf{TBT(ms)}  \\ 
\hline
\multirow{6}{*}{\parbox{1.8cm}{SpecBench\\(Vicuna-7B)}}  & $\times$ & $\times$ & $\times$ & 655.6 & 52.3 \\  
                                                        & $\times$ & \checkmark & $\times$ & 384.0 & 39.4 \\  
                                                        & \checkmark & $\times$ & $\times$ & 655.3 & 38.3 \\
                                                        & \checkmark & $\times$ & \checkmark & 657.9 & 33.8 \\
                                                        & \checkmark & \checkmark & $\times$ & 384.3 & 30.6 \\
                                                        & \checkmark & \checkmark & \checkmark & 384.2 & 26.4 \\ 
\hline 
\multirow{6}{*}{\parbox{1.9cm}{CNN/DM\\(Vicuna-13B)}}      & $\times$ & $\times$ & $\times$ & 1989.0 & 128.1 \\  
                                                        & $\times$ & \checkmark & $\times$ & 1038.7 & 74.2 \\  
                                                        & \checkmark & $\times$ & $\times$ & 1993.5 & 84.3 \\
                                                        & \checkmark & $\times$ & \checkmark & 1992.1 & 74.9 \\
                                                        & \checkmark & \checkmark & $\times$ & 1039.1 & 49.3 \\
                                                        & \checkmark & \checkmark & \checkmark & 1039.9 & 43.5 \\ 
\hline 
\end{tabular}}
\end{table}

There are three key strategies of \method, speculative decoding (SD), prompt chunking (PC) and parallel drafting (PD), being developed to enhance the performance of U-shaped inference.
Herein, we conduct several sets of experiments to evaluate the effectiveness of these critical strategies.
We adopt the U-shape inference with some of these strategies as baselines.
By Table \ref{table:Ablation Study}, SD plays a significant role in reducing the TBT. 
The PC mechanism significantly reduces the TTFT by processing the transmission and computation in parallel.
Besides, the PC mechanism can also reduce the TBT through the chunking operation. 
Furthermore, the PD mechanism further reduces the TBT by paralleling the drafting and verification stages of the SD.
For example, compared to the U-shape inference on SpecBench, the introduction of SD can reduce TBT by 26.8\%. 
When further combined with the PC mechanism, both the TTFT and TBT are reduced by 41.4\% and 41.5\%, respectively.
Furthermore, the PD mechanism can further reduce the TBT by 49.5\%.
Similarly, the TTFT and TBT of U-shape inference on CNN/DM are 1989.0ms and 128.1ms on CNN/DM, respectively. 
The implementation of SD alone reduces the TBT to 84.3ms.
When the PC mechanism is applied independently, it achieves the TTFT of 1038.7ms and the TBT of 74.2ms.
Moreover, when SD and PC are utilized together without PD, \method manages to decrease the TBT further by 61.5\%.
The results reflect the positive roles of speculative decoding, prompt chunking, and parallel drafting mechanisms in \method.

\subsection{Effect of Pipeline Length Scale}\label{Effect of Pipeline Length Scale}

\begin{figure}[!t]
\centering
\subfigure[TTFT]
{
    \includegraphics[width=0.46\linewidth]{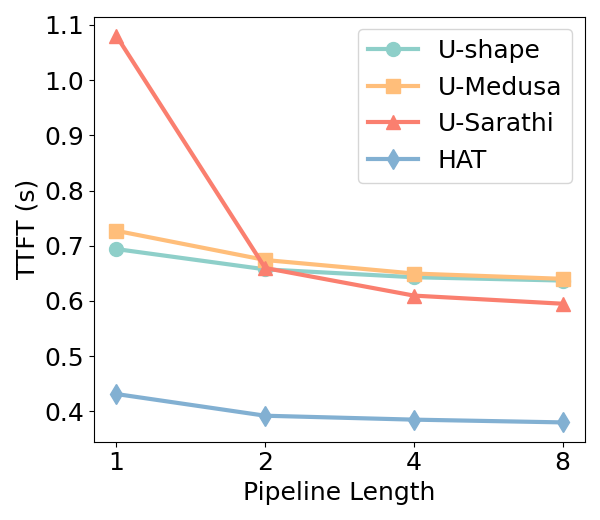}
    \label{fig: SpecBench_pipe_TTFT}
}
\subfigure[TBT]
{
    \includegraphics[width=0.46\linewidth]{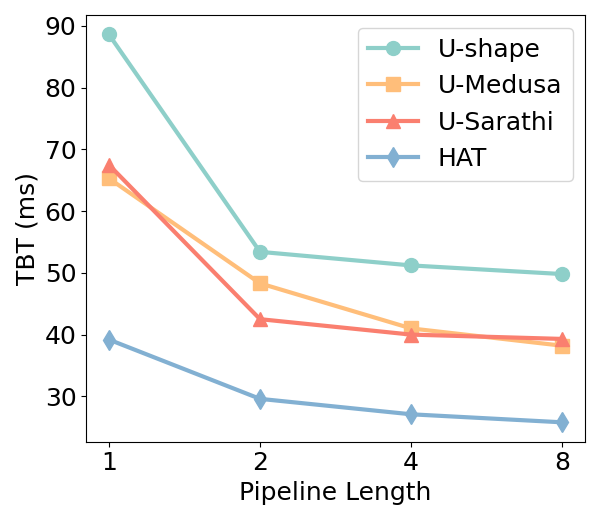}
    \label{fig: SpecBench_pipe_TBT}
}
\vspace{-0.3cm}
\caption{Delay with Different Pipeline Length on SpecBench.}
\label{lab: pipe_specbench}
\end{figure}

\begin{figure}[!t]
\centering
\subfigure[TTFT]
{
    \includegraphics[width=0.46\linewidth]{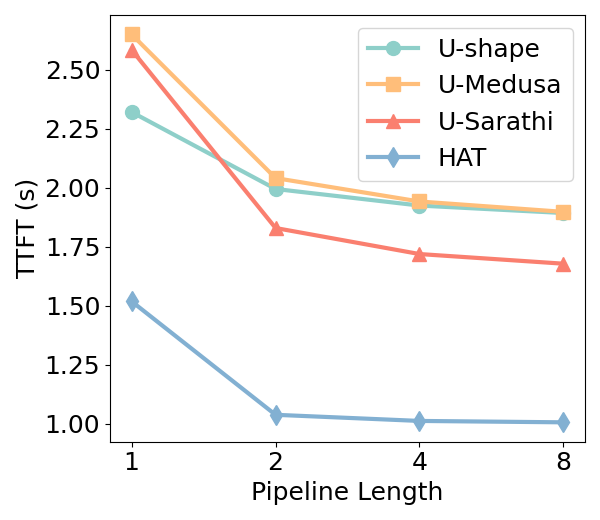}
    \label{fig: CNNDM_pipe_TTFT}
}
\subfigure[TBT]
{
    \includegraphics[width=0.46\linewidth]{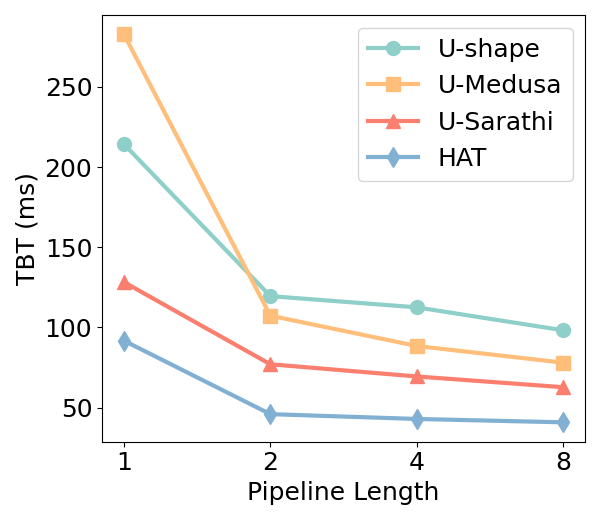}
    \label{fig: CNNDM_pipe_TBT}
}
\vspace{-0.3cm}
\caption{Delay with Different Pipeline Length on CNN/DM.}
\label{lab: pipe_CNNDM}
\end{figure}

To demonstrate the robustness of \method, we evaluate the performance of \method and baselines with different pipeline lengths in the server.
Increasing the server's pipeline length can effectively reduce waiting delay as each GPU has less computation delay.
As shown in Figures \ref{lab: pipe_specbench}-\ref{lab: pipe_CNNDM}, both the TTFT and TBT of all approaches decrease as pipeline length increases.
However, \method consistently outperforms the other approaches.
At a minimal pipeline length of 1, request accumulation within the server leads to increased delay, as incoming requests cannot be processed promptly.
Besides, U-Sarathi increases the number of inference steps in \prefill phase by chunking, leading to high TTFT. 
Specifically, by Figure \ref{lab: pipe_specbench}, with a pipeline length of 1 on SpecBench, \method achieves the TTFT of 431.4ms and the TBT of 39.2ms, while the TTFT for U-Sarathi, U-Medusa, and U-shape measure 1080.0ms, 727.1ms, and 694.0ms respectively, with corresponding TBT of 67.5ms, 65.3ms, and 88.6ms.
Moreover, at a pipeline length of 4 on CNN/DM, Figure \ref{lab: pipe_CNNDM} demonstrates that \method reduces the TTFT by about 36.9\%, 40.7\%, and 40.1\%, while reduces the TBT by about  32.3\%, 33.9\%, and 47.1\%, compared to U-Sarathi, U-Medusa, and U-shape, respectively.
These results further illustrate the robustness of \method.

\section{Related Work}\label{sec:related}
\subsection{Cloud-based LLM Inference} 


Commercial cloud-based LLM systems typically optimize for high throughput by processing multiple concurrent requests from diverse devices. 
To address this requirement, Orca \cite{yu2022orca} introduces continuous batching, which dynamically integrates and removes requests during forward passes without padding or batch completion constraints, significantly improving system throughput. 
Building on this concept, vLLM \cite{kwon2023vllm} combines continuous batching with priority scheduling, particularly prioritizing the \prefill phase.
While effective, this prioritization can interfere with ongoing decoding processes, potentially compromising decoding rates due to the computationally intensive nature of \prefill phase, which introduces delays in \decode phase.
Therefore, Sarathi-Serve \cite{agrawal2024taming} divides the prompt to fully utilize the computational resources and reduce computation delay for \decode phase.
In contrast to these existing approaches, our framework leverages local device computational resources while simultaneously minimizing device-cloud communication latency, offering a distinct advantage in device-cloud collaborative inference scenarios.

\subsection{Speculative Decoding}
Speculative decoding has emerged as a pivotal technique in device-cloud collaborative inference frameworks to address the autoregressive limitations of LLMs. 
For example, Hao \etal \cite{hao2024hybrid} integrate speculative decoding into the device-cloud framework, and systematically analyze how acceptance thresholds influence both inference accuracy and speed. 
Similarly, Oh \etal \cite{oh2024uncertainty} develop a hybrid framework that leverages on-device SLMs and the in-cloud LLM, utilizing the uncertainty of SLM to dynamically adjust LLM acceptance probabilities and verification processes, thus achieving optimal generation speed with negligible accuracy degradation.
To further refine speculative decoding, Medusa \cite{cai2024medusa} introduces a self-drafting mechanism to minimize drafting overhead while employing tree-based verification to maximize accepted sequence lengths.
Concurrently, Eagle \cite{li2024eagle} enhances acceptance probabilities through a specialized autoregressive head trained via feature-layer knowledge distillation. 
However, these methods involve sharing input and output tokens, posing potential privacy risks.


\subsection{Split Inference}
Split inference has emerged as a privacy-preserving framework for device-cloud inference.
For example, Sa \etal \cite{sa2024ensuring} employ U-shaped inference to safeguard both input and output data from exposure beyond devices.
Similarly, Ohta \etal \cite{ohta2023lambda} resides LLM's middle submodel in the cloud, enabling computation offloading while protecting privacy.
Based on split inference, Mai \etal \cite{mai2023split} add noise to intermediate data, enhancing privacy at the cost of some inference accuracy.
Moreover, Shen \etal \cite{shen2023split} divides the LLM and then fine-tunes the on-device submodel  with private data, which meets individual demand while enhancing privacy.
However, these methods do not address the communication delay issue due to the massive and frequent transmission of hidden states.



\section{Conclusions}\label{sec:conclusion}

In this paper, we propose a novel device-cloud collaborative inference framework, named \method, which integrates the advantages of U-shaped inference and speculative decoding to provide LLM services with low delay, high accuracy, and enhanced privacy.
Besides, we propose a prompt chunking mechanism and a parallel drafting mechanism to accelerate inference speed.
The experimental results show that the \method can reduce the TTFT by 41\% to 54\% and the TBT by 41\% to 77\%, compared to the baselines.






\balance


\bibliographystyle{IEEEtran}
\bibliography{USD}

\end{document}